# Shedding Light on Blind Spots – Developing a Reference Architecture to Leverage Video Data for Process Mining

Wolfgang Kratsch,[1,3,*] Fabian König,[1,3] Maximilian Röglinger[2,3]

[1]*FIM Research Center, University of Applied Sciences Augsburg*
[2]*FIM Research Center, University of Bayreuth*
[3]*Project Group Business & Information Systems Engineering of the Fraunhofer FIT*

**Abstract**

Process mining is one of the most active research streams in business process management. In recent years, numerous methods have been proposed for analyzing structured process data. In many cases, however, only the digitized parts of processes are directly captured by process-aware information systems, whereas manual activities often leave blind spots in the process analysis. While video data can contain valuable process-related information that is not captured in information systems, a standardized approach to extracting event logs from unstructured video data remains lacking. To solve this problem and facilitate the systematic usage of video data in process mining, we have designed the ViProMiRA, a reference architecture that bridges the gap between computer vision and process mining. The various evaluation activities in our design science research process ensure that the proposed ViProMiRA allows flexible, use case-driven, and context-specific instantiations. Our results also show that a prototypical implementation of the ViProMiRA is capable of automatically extracting more than 70% of the process-relevant events from a real-world video dataset in a supervised learning scenario.

# 1 Introduction

Process mining strives to discover, monitor, and improve processes by extracting knowledge from event logs commonly available in information systems [1]. Recently, process mining has evolved into one of the most active and fast-growing research streams in business process management (BPM), providing evidence-based decision support for the day-to-day management of processes [2]. The first International Conference on Process Mining hosted in 2019 in Aachen underlines the scientific relevance of the subject [3]. In practice, Celonis' super-fast expansion from start-up to unicorn in only seven years indicates the enormous cross-industry business potential of process mining. By 2023, process mining technologies are predicted to reach a market potential of US$1.42 billion [4].

---

[*] corresponding author

Current process mining applications rely heavily on structured business data, gathered from relational databases of process-aware information systems or other enterprise information systems, such as enterprise resource planning (ERP) systems [5]. However, according to Forbes and CIO magazine, 80 to 90% of available data is unstructured, which is to say it has no retrievable data scheme [6,7]. Moreover, the volume of unstructured data is increasing more rapidly than that of structured data [7]. Since most mature analyses focus on digitized processes recorded in information systems, process mining only covers a proportion of 10 to 20% of the available data. Thus, it is not possible to analyze processes from end-to-end if these processes, for example, involve manual activities that are not tracked in information systems.

In this work, we use the term "blind spot" to refer to parts of real-world processes that cannot be captured in event logs yet because they are not completely enacted in information systems [8]. Filling these blind spots with manual observations can be very time-consuming and is therefore an unscalable option. In many cases, however, vast amounts of unstructured data, such as media files or text documents, are available and may contain valuable information about activities that occur within these blind spots. Consequently, all of the academic experts questioned in a recent Delphi study stated that BPM should prioritize the exploration of unstructured data [9].

There are first approaches that propose techniques to make unstructured data usable for process mining. Some of them apply natural language processing to text documents [10,11]. Others equip actors [12] or things [13] with sensors to generate event logs and apply process mining techniques. Such sensor-based approaches, however, cannot be scaled for use in wider contexts as measured values are dependent on the deployment location. Furthermore, full equipment with sensors appears to be an unrealistic scenario in the case of broad system boundaries or open systems, for instance, when external actors are included. In contrast, techniques based on natural language processing are much easier to generalize, but – just like structured log data – describe only activities performed within information systems (e.g., mail systems). While compliance with privacy regulations must be ensured and privacy aspects must be carefully considered [14], video data from commonly used cameras has the potential to make processes with blind spots more observable.

Initial technology-driven approaches already support the analysis of video data for specific use cases such as object detection and activity recognition in highly specific contexts of production and logistics, often in laboratory settings [15]. The successful application of most recent computer vision approaches, which build on deep learning, suggest that computer vision could be the key to extracting, piece-by-piece, structured information from a vast amount of unstructured video data. Were this to be transferred to process mining, events and actors extracted from video data could be fed into structured event logs that can be processed with various existing process mining algorithms and software.

To date, early research on the intersection of computer vision and process mining has been applied with promising results [16]. To the best of our knowledge, however, there is no tried and trusted set of guidelines on how to holistically integrate computer vision capabilities into process mining. Thus, our research question is: *How can video data be systematically exploited to support process mining?*

We answer this research question by proposing the Video Process Mining Reference Architecture (ViProMiRA). The artifact makes it possible to extract structured process information from unstructured video data and transform it into a format suitable for process mining. To this end, ViProMiRA facilitates the use case-driven implementation and integration of computer vision capabilities into process mining. As a general research process, we adopt the design science paradigm [17], and perform multiple evaluation activities to show the usefulness and real-world applicability of our artifact. By instantiating the ViProMiRA as software prototype and applying it to a real-world dataset, we illustrate which computer vision capabilities are suitable for which process mining contexts.

Our study is organized as follows: Section 2 provides the theoretical background on process mining and computer vision since both are integral to ViProMiRA. Section 3 presents the applied research design. The remainder of this paper is structured according to the design science research (DSR) reference process as per Peffers et al. [18], which necessitates the justification of the research gap and presentation of related approaches as a first phase in Section 4. Section 5 outlines the design objectives (DOs), to which any solution for systematically exploiting video data in process mining must conform. Section 6 describes the ViProMiRA, which is then instantiated as a software prototype in Section 7, and applied to a real-world dataset in Section 8. Section 9 concludes the paper with a look at the limitations and a perspective on future research.

## 2 Theoretical Background

### 2.1 Process Mining Use Cases

Initially, the process mining manifesto defined three use cases, namely process discovery (generation of as-is models), conformance checking (comparing as-is with to-be models), and model enhancement (improve existing models through log insights) [1,19]. Generally, process mining use cases can be structured around their required data sources (i.e., current or historical event data as well as normative or as-is process models) [20]. Accordingly, all use cases taking historical data to analyze the past can be subsumed as ex-post use cases. These ex-post use cases mainly relate to cartography (i.e., use cases that work with process models or maps) and auditing (i.e., use cases that check whether business processes have been executed within certain boundaries set by stakeholders). Some auditing use cases also take normative data (i.e., de jure models) to explore deviance in historical cases (i.e., post-mortem data). All ex-post approaches are descriptive in nature, which means that they analyze and visualize what happened in the past without looking into the future [20].

In recent years, the scope of process mining has evolved from predominantly backward-looking analysis to forward-looking operational decision support [2]. So-called ex-ante use cases are increasingly becoming the focus of research. Ex-ante use cases are based on historical data and, in some cases, on partial trace information from ongoing cases (i.e., pre-mortem data). Predictive process monitoring, for example, applies predictive models to correlate extracted features from partial trace information with historical traces in real-time [21]. Predictive process monitoring approaches differ in terms of applied methods and, more importantly, the target of prediction [22]. Such targets can be the remaining cycle time of an ongoing case [23], the outcome or an anomaly of a case [24], or the next action that will take place in further case processing [25]. The latter prediction task reveals that there is a fluid transition between prediction and prescription. By predicting the next action and identifying the decision area from normative de jure process data, one can set up models that predict the outcome of each possible action and recommend the most favorable one [20]. The transition to prescriptive process analytics is achieved when forward-looking guidance is automatically provided on how to meet a specific objective [26].

## 2.2 Event Log Extraction and Event Abstraction in Process Mining

Organizations store process-related data in highly distributed ways, e.g., structured data in relational databases and unstructured data in object databases, or even in big data architectures such as data lakes. As process mining requires a highly standardized flat event log, references to events, cases (i.e., process instances), and related attributes such as timestamps must be extracted and converted from underlying source data [27]. The IEEE Task Force on Process Mining defined eXtensible Event Stream (XES) as a standardized format for storing event logs [28]. XES structures events along traces, allowing the correlation of events to cases. Being a denormalized, flat data structure, traces serve as containers storing all related structured attributes (e.g., Strings, Numbers, or Booleans).

Extracting XES event logs is one of the key challenges in process mining [1]. While most research conducted in this area has focused on the use of new mining techniques applied to synthetically generated event logs or the few event logs published for research (e.g., Business Process Intelligence Challenge event logs [29]), only a handful of studies address the challenge of extracting real-world data for process mining [30]. Andrews et al. [5] list several approaches targeting structured data sources, such as ERP systems, other process-aware information systems, and – more generally – relational databases.

Since information systems only cover a small fraction of the real world, it is reasonable to extend event logs with supplementary data sources, some of which might be unstructured [31]. As discussed in Section 4, only a few approaches have been proposed that target unstructured data for event extraction. Furthermore, existing video-based approaches do not propose a sufficiently general answer as to how event logs can be extracted from video data. To obtain meaningful high-level business events from raw data, it is necessary to apply event abstraction concepts that preprocess and filter recorded low-level events [32]. For a comprehensive overview of event abstraction concepts, we refer to van Zelst et al. [33]. As for the nature of data, one can distinguish between concepts that assume discrete simple events and others that can handle continuous event streams, e.g., complex event processing [33]. Complex event processing obtains relevant situational knowledge from low-level event streams in real-time by selecting, aggregating, and abstracting events to generate higher-level complex events of interest [16,34].

## 2.3 Computer Vision for Information Extraction

Computer vision applies mathematical techniques to visual data (e.g., images and videos), striving to achieve or even surpass human-like perceptual interpretation capabilities [35–37]. To date, computer vision has enabled several real-world use cases, including self-driving cars, facial recognition [38], and the analysis of medical images for healthcare [39]. Although various computer vision capabilities exist, this study focuses on capabilities that extract structured information from visual data (i.e., conventional RGB data) rather than modifying the visual data itself (e.g., producing three-dimensional shapes from two-dimensional image data [40]). This focus makes it possible to identify computer vision capabilities that can extract low-level events from video data, which, as described in Section 2.2, serve as a basis for obtaining meaningful high-level business events for process mining.

An overview of approaches that implement and combine computer vision capabilities can be found in the supplementary material (Table A.1).[1] Based on these approaches and a review of general research in computer vision, we identified nine relevant computer vision capabilities [41–43], which, with the exception of background subtraction and re-identification, are exemplified in Figure 1. In the following, we will briefly describe all computer vision capabilities and sketch their purpose.

- **Background Subtraction** enables the detection of non-static information (e.g., moving objects) from video data recorded with a static perspective by computing the difference between any frame and a predefined reference frame that contains the normal scene information [44].
- **Image Classification** predicts probabilities for the occurrence of certain object classes (e.g., people) in images [42].

---

[1] https://github.com/FabiTheGabi/VideoProcessMining/blob/master/Supplementary%20Material.pdf

accepted manuscript at Decision Support Systems        6

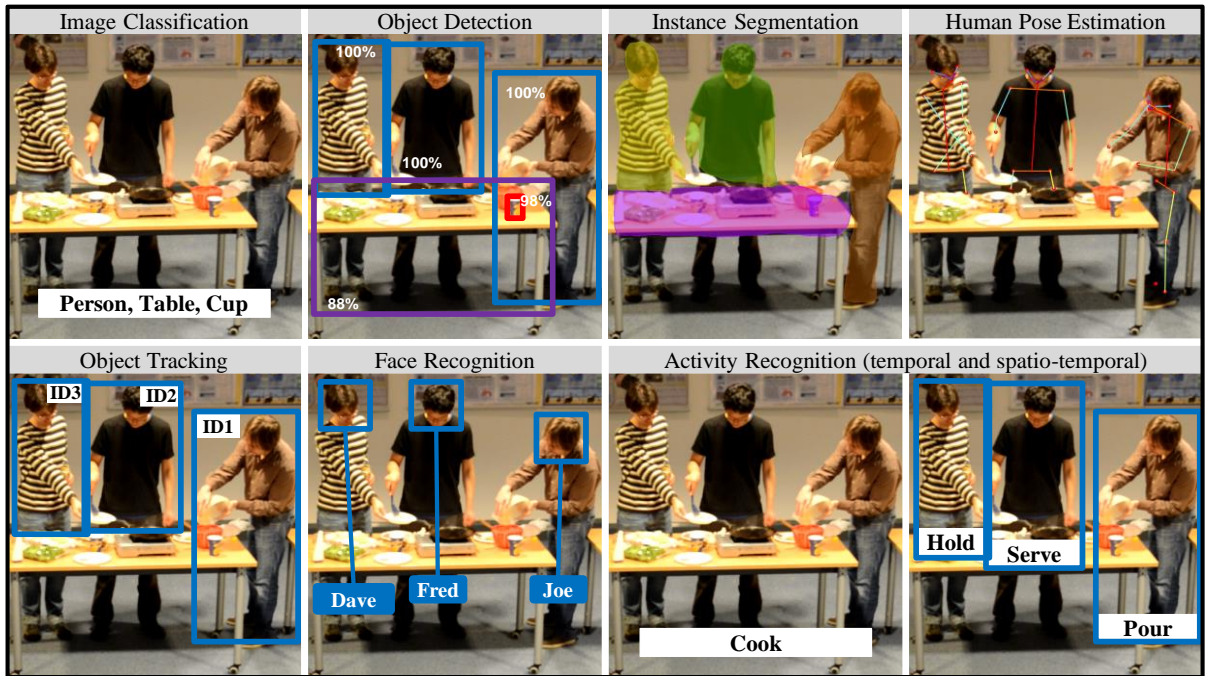

Figure 1: Illustration of computer vision capabilities with an image of the dataset from [45]

- **Object Detection** finds instances of object classes and further localizes their positions [41,42]. The position information is often represented by rectangles (i.e., bounding boxes).

- **Semantic Segmentation** labels all pixels with their enclosing object class, while **Instance Segmentation** additionally distinguishes between instances of the same classes [46].

- **Human Pose Estimation** addresses the problem of localizing human body parts or anatomical key points (e.g., elbow, wrist) in images [47,48].

- **Object Tracking** predicts the trajectory of a target object along a sequence of frames and assigns a stable track identifier [49,50]. Multiple object tracking extends this original goal by simultaneously tracking several instances of at least one object class [51]. **Re-identification** is a sub-capability that makes it possible to identify a person of interest across multiple cameras [52].

- **Face Recognition** (re)identifies individuals by their faces, which can be a challenging endeavor due to head rotation, facial expression, or aging [38,53].

- **Activity Recognition** takes the output of other computer vision capabilities (e.g., object detection or human pose estimation) to identify the activities of at least one person in a sequence of image frames (i.e., video) [54,55]. In temporal activity recognition, which bears a resemblance to image classification, an entire video sequence is assigned to one or several activity classes. Spatio-temporal activity recognition extracts several concurrent activities (e.g., the movements of two tennis players) in the same video sequence. This is more challenging as it provides the location of all activities as well as their start and end times [56].

## 3 Research Design

To answer our research question, we constructed the ViProMiRA as artifact to systematically bridge the gap between computer vision and process mining. Since the ViProMiRA makes statements about relationships among constructs in the sense of a model, it is a valid design artifact [57]. Thus, we adopted the DSR paradigm [17] and followed the DSR reference process proposed by Peffers et al. [18], which includes six phases. As the generic DSR process does not provide any guidance on how to design reference architectures, it needs to be complemented by a research method that is appropriate for this artifact type. Therefore, we adopted the method presented by Galster and Avgeriou [58], which specifically aims at building reference architectures. Since this stand-alone method partially overlaps with the DSR process, we only performed the steps that focused on the design of the reference architecture and were not covered by the generic DSR process. Moreover, we conducted multiple evaluation activities (i.e., EVAL1-EVAL4) throughout the research process [59] that allowed us to continuously assess our artifact. The evaluation was guided by established criteria (i.e., real-world fidelity, understandability, internal consistency, applicability, and usefulness) [57,59]. Figure 2 illustrates our end-to-end research design. Below, we outline how we performed the DSR process, which steps we selected from Galster and Avgeriou [58], and which evaluation activities we conducted.

In Phase 1, the *Problem Identification & Motivation* phase (Section 4), we conducted a semi-structured literature search to demonstrate that the research problem articulated in Section 1 cannot be solved by existing approaches (EVAL1). In Phase 2, *Definition of Design Objectives* (Section 5), we derived DOs from the literature. DOs describe "how a new artifact is expected to support solutions to problems not hitherto addressed" [18]. In Phase 3, the *Design & Development* phase (Section 6), we created the ViProMiRA. To complement the DSR process, we followed the method presented by Galster and Avgeriou [58], which consists of six steps: (I) decision on the type of reference architecture, (II) selection of the design strategy, (III) empirical acquisition of data, (IV) construction of the reference architecture,

(V) enabling the reference architecture with variability, and (VI) evaluation of the reference architecture. Since the evaluation in step (VI) is covered by the DSR process, we only performed steps (I) to (V).

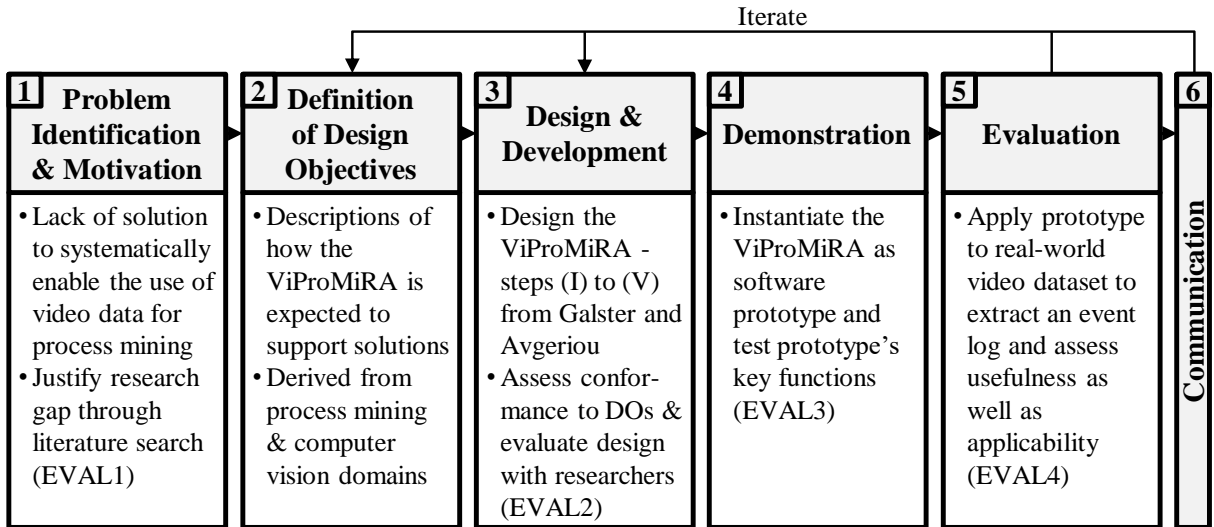

Figure 2: Design science research process based on Peffers et al. [18]

For the type of reference architecture (I), we chose a preliminary reference architecture that facilitates the design and implementation of video-based process mining in various domains as it was constructed by independent researchers (variant 5.1 in [60]). To ensure that the design strategy (II) corresponds to this preliminary nature of the ViProMiRA, we built our artifact from scratch using related research as a basis. As for empirical data acquisition (III), we collected relevant research on different process mining use cases and computer vision capabilities (Section 2). We constructed the ViProMiRA (IV) based on the acquired information. As we strived to develop a preliminary reference architecture that also provides operational support for real-world implementation, we chose a semi-concrete, semi-detailed UML-based documentation level [58]. To ensure appropriate variability of the ViProMiRA (V), we embedded variation points, which should be aligned to the targeted process mining use case, the process context, and the degree of desired generalization capability. The evaluation of the ViProMiRA's design specification (EVAL2) was twofold. First, we applied logical reasoning to show that the ViProMiRA provides a solution to the research problem in terms of complying with the DOs. Then, to determine whether the ViProMiRA adequately combines relevant concepts from the domains of process mining and computer vision, we assessed its real-world fidelity and understandability with fellow researchers. In Phase 4, the *Demonstration* phase (Section 7), we instantiated our proposed ViProMiRA as software prototype and repeatedly tested its key functions to validate the internal consistency and applicability of our artifact (EVAL3). In Phase 5, the *Evaluation* phase (Section 8), we assessed the artifact's usefulness and applicability by applying our prototype to a publicly available real-world video dataset. We used the prototype to extract an event log, which we analyzed with process mining software (EVAL4). Finally, in Phase 6, *Communication*, we present our research in its published form.

## 4 Problem Identification (EVAL1)

To ensure that the research problem presented in Section 1 cannot be solved by existing approaches, we conducted a semi-structured literature search [59]. We referred to the taxonomy on sensor-based activity recognition and process discovery in industrial environments by Mannhardt et al. [61] as a starting point and did a forward and backward search on process mining research that focuses on event abstraction and video data. As there is hardly any research in our addressed field, we also searched for approaches that offer partial solutions or solve similar research problems.

Table 1 provides an overview of the approaches identified, which are summarized in the supplementary material (Table B.1). While some approaches enable users to extract events from unstructured documents [62–64], others use sensor data to perform activity monitoring in manual production processes and aggregate extracted events [12,13,65]. However, most of the presented approaches are only distantly related to process mining, as they either do not provide the extracted event log (e.g., [12]) or address only a discovery use case (e.g., [65]).

Table 1: Approaches for event extraction from unstructured data sources

| Reference | Type of Data | Process Mining Use Cases | Level of Abstraction |
|---|---|---|---|
| Pospiech et al. [62],[63] | Text | Discovery | Semi-concrete, high-level architecture |
| Yang et al. [64] | Text | Discovery, exploration | Semi-concrete, domain-specific architecture |
| van Eck et al. [13] | Sensor | Discovery, conformance | Concrete, case study |
| Cameranesi et al. [65] | Sensor | Discovery | Concrete, case study |
| Raso et al. [12] | Sensor | Discovery, conformance | Semi-concrete, experimental setting with specific architecture |
| Knoch et al. [16] | Sensor, video | Discovery, conformance, enhancement | Semi-concrete, laboratory experiment with generalized architecture |
| Knoch et al. [66] | Video | Conformance, enhancement | Concrete, case study (building on [16]) |
| Knoch et al. [67] | Video | Discovery, conformance | Concrete, laboratory case study with a similar setup as in [16,66] |
| Rebmann et al. [68] | Sensor, image | Discovery | Semi-concrete, laboratory experiment with generalized architecture |

Since existing approaches apply video data mainly to address domain-specific research problems, they were not originally designed to answer our research question. Nevertheless, to ascertain whether existing research could solve the problem addressed in this work, we analyzed the most promising approaches in detail. To qualify for this close analysis, the approaches from Table 1 had to meet the following three criteria: they proposed a generic solution for event extraction from unstructured data, addressed at least one process mining use case, and included an implementation or instantiation to demonstrate the applicability of the solution presented. After applying the selection criteria, five research items remained (i.e., [12,16,66–68]). Since the case studies in [66] and [67] extend the work presented in [16], we considered them together. While all of the approaches address at least one ex-post process mining use case, they do not target ex-ante process mining use cases at all. Three approaches [16,67,68] generate event logs, but only [67] provides the logs in the standardized XES format. Furthermore, only [16,67] illustrate the aggregation of raw data into high-level events. With the exception of [66,67], which primarily use video data as input, the approaches explore settings with multiple sensors. Although [68] apply image classification, and [16,66,67] additionally incorporate background subtraction and object detection, no approach introduces other suitable capabilities in in a way that could be generalized.

As is apparent from EVAL1, existing approaches provide valuable results in their respective area of research, but they are not suited to systematically exploit the potential of video data for process mining. Consequently, the development of a novel artifact, which extends the knowledge base, is required.

## 5  Definition of Design Objectives

In the second phase of our DSR process, outlined in Figure 2, we defined DOs that an artifact should achieve to provide a solution to systematically exploit video data in process mining [18]. To purposefully guide the development and the evaluation process of the ViProMiRA, we derived the DOs described in Table 2 from the relevant literature in the fields of process mining and computer vision, as discussed in Section 2.

Table 2: Design objectives for the ViProMiRA

| | DO | Description |
|---|---|---|
| **DO1: Use case orientation** | DO 1.1 | *The artifact must support ex-ante and ex-post process mining use cases:* Process mining has evolved from backward-looking descriptive process analytics into a forward-looking ex-ante mode [20]. However, initial ex-post use cases, such as discovery, still serve as a starting point for more sophisticated process mining use cases [2]. |
| | DO 1.2 | *The artifact must be flexible to support the targeted process mining use case efficiently:* Several use cases require a specific extraction of information [1]. For instance, anomaly detection may not even require the output of a structured log, whereas conformance checking requires mapped activity names and cases correlated to recorded events. |
| **DO2: Event log extraction** | DO 2.1 | *The artifact must consider the extraction of XES-conforming event logs from video data:* XES is a well-accepted standard, ensuring a consistent and accurate data model [28]. As most available process mining tools natively support XES, the exported event logs can be analyzed with established process mining software. |
| | DO 2.2 | *The artifact must comprise abstraction and generalization capabilities to extract meaningful high-level events or activities from an unstructured video event stream:* As video data represents unstructured data, it is necessary to apply event abstraction and generalization concepts to get from low-level events to insights at the process level [33]. |
| **DO3: Computer vision capabilities** | DO 3.1 | *The artifact must cover relevant computer vision capabilities:* The computer vision domain offers a rich body of knowledge on specialized methods to extract structured information from video data [37]. This knowledge should be considered when designing the artifact. |
| | DO 3.2 | *The artifact must guide the selection of appropriate computer vision capabilities for different process mining use cases and contexts:* A specific process mining context requires a tailored selection of computer vision capabilities, determined not only by the respective process mining use cases but also by the characteristics of the process space (e.g., number of actors, view of the camera). |
| | DO 3.3 | *The artifact must provide guidance on how to put different computer vision capabilities into practice:* Typically, several computer vision capabilities are combined to extract different aspects of a video sequence. The artifact should guide users through the thicket of potential combinations. |

## 6 Design and Development

### 6.1 Design Specification of the ViProMiRA

Figure 3 illustrates the ViProMiRA, including its three subsystem layers of *Data Preprocessor*, *Information Extractor*, and *Event Processor*. The ViProMiRA connects to various BPM applications and, thus, supports diverse process mining use cases based on video data. To this end, it provides guidance in the form of colored instantiation variants on how to integrate relevant computer vision capabilities to systematically exploit the potential of video data in process mining. To make sure the ViProMiRA can also be adapted to specific process mining use cases, it further comprises a range of optional components, as indicated by the dotted frames.

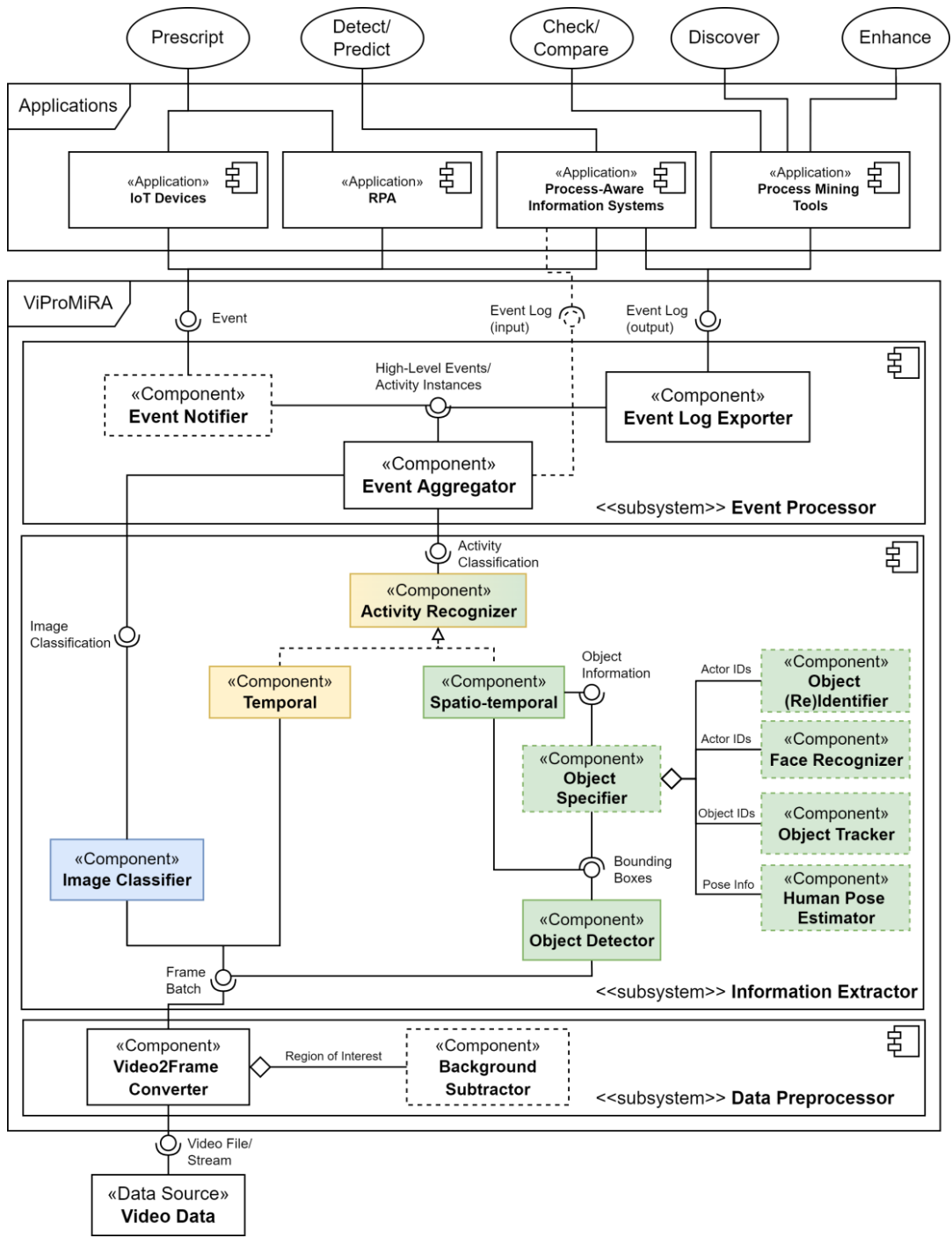

**Figure 3: The ViProMiRA as UML diagram**
**Dotted frames indicate optional components. Colors indicate instantiation variants.**

The **Data Preprocessor** serves as an interface to the raw video file or stream. During preprocessing, the *Video2Frame Converter* synchronizes different frame rates, converts continuous video streams into single frames, and adjusts their resolution. The constant frame rate across different input videos guarantees consistent and comparable time intervals of frame sequences, which is a prerequisite for any meaningful activity recognition. In static camera settings, the *Background Subtractor* can be applied to extract certain regions of interest. To comply with legal regulations, such as Europe's General Data Protection

Regulation, the subjects in view must have explicitly consented to the processing of their data [14]. A further requirement is transparency about who uses and processes the information, and the purpose for which this is done. If the explicit consent of the subjects cannot be collected, sensors must be used that comply with data protection laws, such as blind sensors (e.g., [69]).

The **Information Extractor** receives the preprocessed frame sequences and combines different computer vision capabilities to extract meaningful information in a hierarchical manner. All components located in the Information Extractor are trainable as they rely on machine learning algorithms. Alternatively, pre-trained models that offer out-of-the-box functionality should be used for standard tasks. The simplest, *blue instantiation variant* consists of the *Image Classifier* as the only component. Its purpose is to extract simple yet potentially important information by classifying single frames (e.g., differentiating between simple categories such as "normal" or "abnormal"). For more sophisticated information extraction, an *Activity Recognizer* is required. There are two possible instantiation variants. The *Temporal Activity Recognizer* is part of the *orange variant* and designed to target process scenarios without concurrency (i.e., activities that are sequentially performed by a single actor). It returns prediction scores for all predicted activity classes, based on a frame sequence (i.e., batched frames). The *green instantiation variant* combines different computer vision capabilities and is designed for scenarios with multiple resources that concurrently execute individual activities. The *Spatio-temporal Activity Recognizer* receives the frame sequence as well as the location information of all resources and returns their prediction scores for all activity classes. Consequently, spatio-temporal activity recognition requires the implementation of an *Object Detector* that detects objects in the input frames and provides the respective bounding boxes. Components for semantic segmentation or instance segmentation that allow for a more precise spatial localization of objects are not included in the green instantiation variant. The marginal utility of this more accurate information compared to the Object Detector is limited and does not correspond to the targeted principle of abstraction. To provide additional information on detected objects, which is required to analyze complex processes, the *Object Specifier* consists of four sub-components. The *Human Pose Estimator* locates anatomical key points. Since it can determine the direction of a movement, it is used to improve the accuracy of activity recognition. The *Object Tracker* traces moving objects from frame to frame and assigns identifiers to each tracked object. It is used for the short-term distinction of resources and applied in scenarios with multiple unknown and frequently changing actors (e.g., warehouse loading processes). The Object Tracker performs best with a high input frame rate and requires constant object detection at the cost of higher computational complexity. The accuracy of the *Face Recognizer*, which reidentifies known resources by their faces, is not affected by the input frame rate, making it less resource-intensive. Since the Face Recognizer requires an initial training before it can consistently distinguish and identify resources, it is applied to assign stable identifiers in process contexts with known actors. As the most sophisticated sub-component that aggregates the input of the other sub-components, the *Object (Re)Identifier* is applied to (re)identify multiple known and unknown process actors.

Choosing the appropriate instantiation variant of the Information Extractor depends on the process mining use case. For instance, simple detection or prediction use cases that focus on atypical events (e.g., anomaly detection) are implemented with the blue or the orange instantiation variant. In contrast, the green instantiation variant is selected for a use case, in which a process with multiple actors is to be captured in full. If the perspective shifts or multiple cameras record the footage, the Object Specifier must be used to distinguish and re-identify temporarily occluded actors.

The **Event Processor** takes the low-level information from the Information Extractor, aggregates it to the level of detail required for the process mining use cases (e.g., high-level business events), and provides interfaces for several BPM applications. Figure 4 illustrates how high-level business events can be abstracted from video data. The event abstraction starts with optical sensors in the physical world (i.e., Data Preprocessor subsystem), from which meaningful information in the form of low-level events (e.g., objects) must be distilled (i.e., Information Extractor subsystem). The *Event Aggregator* receives these low-level events, excludes all irrelevant ones, such as macro-changes in the environment or sensor noise, and uses concepts like complex event processing to derive a meaningful, high-level business event stream, for instance, the beginning and end of a worker's movement [34]. If existing structured event logs exist, the Event Aggregator can enrich them with these extracted high-level events. Finally, to derive activity instances, the Event Aggregator associates high-level events with predefined process activities. To do so, the required level of event abstraction must be aligned with the respective process mining use case. Since the high-level events are already well-structured, their contextualization to activity instances may also be outsourced to process mining tools. As depicted in Figure 4, the correlation of activity instances to cases lies beyond the focus of the ViProMiRA. Although typically there is no case identifier assigned to raw streaming data, existing process mining research has already provided solutions for solving the problem of event case correlation (e.g., [70]).

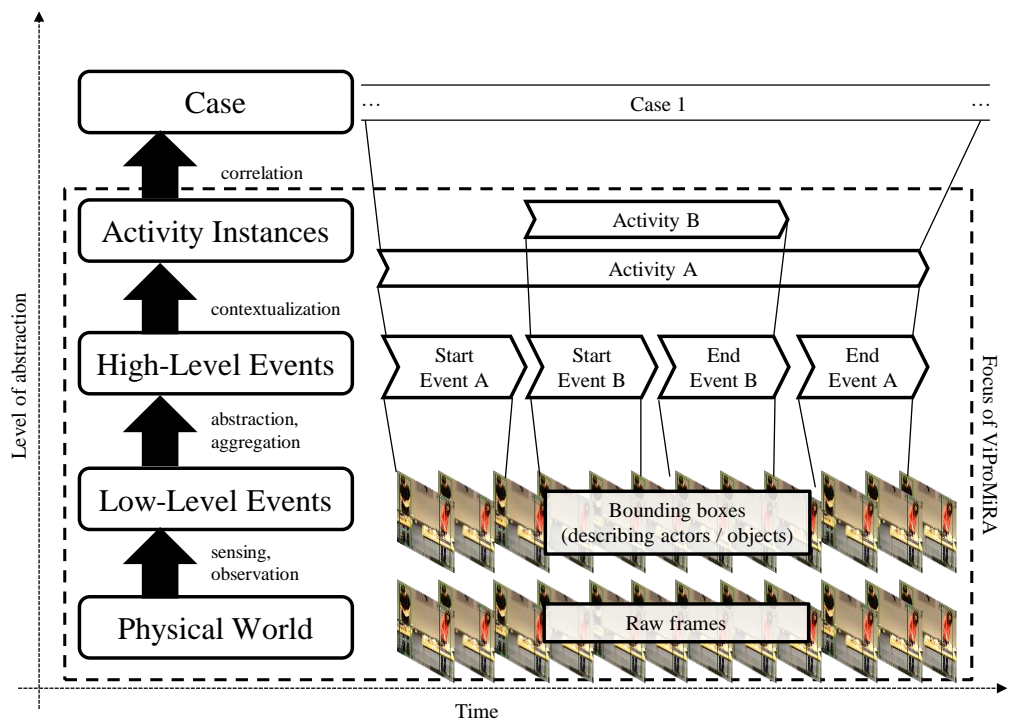

**Figure 4: The concept of event abstraction by the Event Aggregator (based on Koschmider et al. [32])**

Low-level events are sufficiently accurate for some predictive process mining use cases, such as anomaly detection (e.g., an activity category or classified frame per second). More explorative use cases like performance analysis, however, require high-level business events. The *Event Log Exporter* and the *Event Notifier* provide interfaces to several information systems and related applications. To support ex-post process mining use cases that mainly draw on historic data (e.g., discovery, conformance, and enhancement), the Event Log Exporter transforms all aggregated high-level business events into a process-mining-compatible format. The Event Notifier forwards single event notifications, which is sufficient for ex-ante process mining use cases that do not rely on complete event logs. Producing single event notifications facilitates the definition of system interfaces, which can be economically beneficial. However, this requires a more targeted and robust event abstraction in the Event Aggregator. It is also worth noting that raised event notifications become less traceable if event logs are not provided.

## 6.2 Evaluation of the Design Specification (EVAL2)

To show that the design specification of the ViProMiRA conforms to the DOs presented in Section 5 and in doing so represents a valid solution to the research problem, we used logical reasoning [59]. Table 3 contains the joint assessments of all co-authors in the form of condensed qualitative statements about the conformance of the ViProMiRA to the DOs.

Table 3: Conformance of the ViProMiRA to the design objectives

| DO | Justification for Conformance |
|---|---|
| DO 1.1 | The ViProMiRA considers interfaces with several BPM applications and matches them to ex-post and ex-ante process mining use cases. |
| DO 1.2 | The ViProMiRA defines mandatory and optional components (e.g., Event Notifier) that enable use case-dependent configurations and appropriate instantiations. |
| DO 2.1 | The ViProMiRA supports XES-conforming event log extraction (i.e., Event Log Exporter). |
| DO 2.2 | As illustrated in Figure 3 and Figure 4, the ViProMiRA contains the Event Aggregator component that enables complex event processing. |
| DO 3.1 | The Information Extractor subsystem of the ViProMiRA comprises multiple relevant computer vision capabilities based on computer vision literature. |
| DO 3.2 | The ViProMiRA accommodates instantiation variants that provide guidance on which computer vision capabilities to use in a given process mining context. |
| DO 3.3 | The Information Extractor subsystem of the ViProMiRA indicates how computer vision components should build upon each other. |

We further discussed the ViProMiRA with fellow BPM researchers to assess its real-world fidelity and understandability [57]. After presenting and discussing the ViProMiRA, we collected feedback through a short survey [71] that ten researchers answered in full. As shown in Table 4, respondents could anonymously rate the artifact's real-world fidelity and understandability on a seven-point Likert scale [72], with anchors ranging from 1 (strongly disagree) to 7 (strongly agree). Consistent with the qualitative feedback from the discussion, the real-world fidelity and understandability of the ViProMiRA's design specification were rated positively by all ten respondents.

Based on EVAL2, we conclude that the ViProMiRA represents a valid solution to all formulated DOs and can provide an essential contribution for leveraging unstructured video data in process mining.

Table 4: Evaluation criteria rated by ten researchers on a seven-point Likert scale

| Criterion | Minimum | Maximum | Mean | Standard Deviation |
|---|---|---|---|---|
| Understandability | 5.0 | 7.0 | 6.3 | 0.64 |
| Real-world fidelity | 5.0 | 7.0 | 5.9 | 0.70 |

## 7 Demonstration (EVAL3)

To demonstrate the applicability and internal consistency of the ViProMiRA [57], we implemented its most extensive green instantiation variant (Figure 3) as a software prototype. The prototype extracts business events from raw video data and creates standardized XES event logs that facilitate the seamless integration into existing process mining workflows (e.g., using ProM or PM4Py). It features spatio-temporal activity recognition and targets manual processes conducted by at least one human resource. To distinguish different resources, the prototype also implements the Object Tracker component.

The prototype uses both the Python programming language and the PyTorch machine learning library as well as open-source computer vision frameworks such as SlowFast [73] and Detectron2 [74]. A detailed description of how the prototype's components were instantiated can be found in the supplementary material C. Suffice it to say here that the prototype performs activity recognition at one-second intervals, which is also the required annotation format for the supervised training of its Activity Recognizer. In the exported event logs, each activity instance consists of exactly one start and completion event. As explained in Section 6.1, the correlation of activity instances to cases lies beyond the focus of the ViProMiRA. Therefore, each event log extracted by the prototype consists of one default trace. We ran multiple test cycles that allowed us to ascertain the prototype's internal consistency and applicability. Our software repository[2] also comprises a demo video that illustrates how the prototype extracts information and further animates the discovered activities by means of Disco process mining [75].

## 8 Evaluation (EVAL4)

### 8.1 Application of the Prototype to a Real-World Dataset

During EVAL4, we applied our prototype to extract an event log from a real-world video dataset. This allowed a holistic assessment of our artifact's usefulness and real-world applicability in subsequent process mining analyses. The Crêpe Dataset is publicly available and contains 90 minutes of video data with a resolution of 1920 x 1080 pixels in 12 video files [45,76]. Unlike most datasets for activity recognition, it provides spatio-temporal and frame-based labeling of continuous video data with multiple activities, resources, and concurrent cases. The Crêpe Dataset simulates a simplified kitchen scenario, in

---

[2] The repository is publicly available under https://github.com/FabiTheGabi/VideoProcessMining

accepted manuscript at Decision Support Systems 18

which up to three chefs (i.e., resources) individually prepare crêpes and several waiters carry out process-unrelated tasks (e.g., cleaning or fetching a dish). Figure 5 depicts the crêpe preparation process model including the six process variants: 1) lemon sugar, 2) banana chocolate, 3) cheese ham, 4) cheese ham parsley, 5) goat's cheese spinach, and 6) goat's cheese spinach nutmeg. Each case consists of the preparation of a single crêpe by exactly one resource. The Crêpe Dataset comprises nine activity classes (i.e., "cut," "flip," "fold," "grate," "pour," "spread," "sprinkle," "stir," and "transfer") and 53 cases in total.

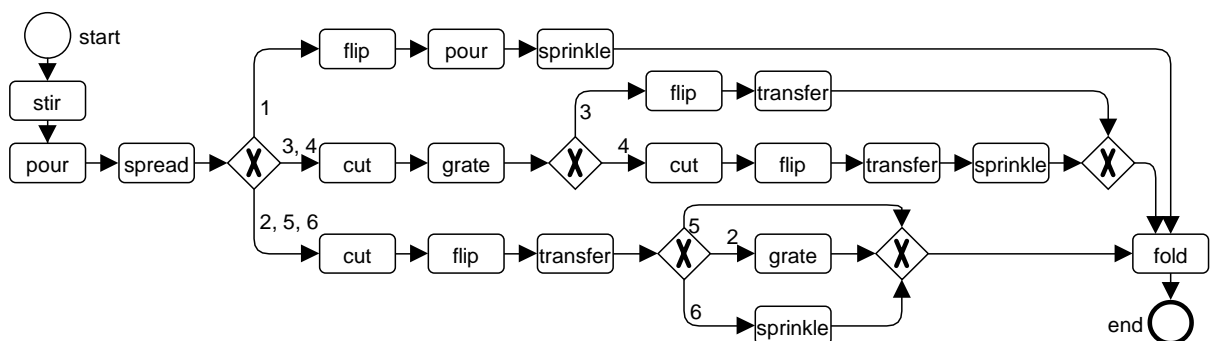

**Figure 5: Crêpe preparation process as process diagram (numbers indicate process variant)**

As a necessary preprocessing step for training the prototype's Activity Recognizer component, we automatically created labels at one-second intervals by using the original frame-based video annotations of the Crêpe Dataset and the videos' frame rates. To enable the distinction between relevant and process-unrelated activities (e.g., chefs waiting or waiters cleaning), we assigned the class "undefined" to all activities without direct impact on the process.

After preprocessing, we split the dataset into 41 cases for the supervised training of the prototype's Activity Recognizer (i.e., ten video files comprising 3308 process-related samples) and 12 cases for evaluation (i.e., two video files comprising 1070 process-related samples). The latter 12 cases were deemed particularly suitable for evaluation because they comprised five different resources, and each process variant was included at least once. To account for the limited amount of training data available for the Activity Recognizer, we applied transfer learning with a pre-trained deep learning model of the Kinetics-400 dataset as a base [77]. Once the training was complete, we used the prototype to extract an event log from the 12 evaluation cases (Extracted_Log). To provide an anchor point for the analysis of the Extracted_Log, we used the preprocessed annotations to generate an event log that included the observable events in the video data (True_Log). Both event logs are described in the following:

- **Extracted_Log**: Represents the events extracted by the prototype. We removed all occurrences of the activity class "undefined", leaving only the nine original process-related activity classes. The Extracted_Log comprises 136 activity instances (i.e., 272 events). As described in Section 7, each event log extracted by the prototype initially stores the information about all cases in one default trace. By considering that only one resource performs a case and that each case starts with the activity "stir," we were able to determine the corresponding trace for each case.
- **True_Log:** Represents the observable events as annotated in the video data. After removing all occurrences of the activity class "undefined," 114 activity instances (i.e., 228 events) remained.

## 8.2 Analysis of the Extracted Event Log

As the output of the prototype, the Extracted_Log was a suitable lens through which to determine how well the individual components of the prototype interoperated. Thus, before performing conformance checking (Section 8.3) and process discovery (Section 8.4) based on the Extracted_Log, we assessed how well it complied with the truly observable behavior. We focused on the activity instance level in the event log and analyzed the following questions: 1) what proportion of all observable behavior was correctly detected (i.e., recall) and 2) what proportion of all detected behavior was correct (i.e., precision) [78]. To answer both questions, we carefully compared the Extracted_Log with the True_Log.

In our analysis, we checked whether an observable activity instance in the True_Log was matched by a detected activity instance in the Extracted_Log, the caveat being that the detection had to lie within the temporal boundaries of the observable activity instance. Therefore, an activity instance in the True_Log could have several matching activity instances in the Extracted_Log, whereas an activity instance in the Extracted_Log could have a maximum of one matching activity instance in the True_Log. Only matching activity instances with the same activity class were counted as a correct match (e.g., activity class "flip" for "flip"). Furthermore, we considered every activity instance that the prototype did not observe although it was contained in the True_Log as "not observed." An activity instance that was incorrectly detected by the prototype and was therefore only present in the Extracted_Log was classified as "not existing" in the True_Log. In cases of discrepancies between detections and observable behavior, we manually examined the evaluation video to ensure a correct assessment.

Due to the instantiation of its Event Processor subsystem, the prototype tended to split extended activities into multiple sub-activities of the same class, which explains the larger number of events in the Extracted_Log. As our assessment logic would have favored these sub-activities, we aggregated the Extracted_Log prior to our comparison: We merged all subsequent events belonging to the same activity class into a single event pair that matched an activity instance in the True_Log. As shown in Table 5, this reduced the number of activity instances in the Extracted_Log to 107 (i.e., 214 events).

Table 5: Overview of event log properties and modifications prior to the analysis

| Name of Event Log | Before Modification | Modification | After Modification |
|---|---|---|---|
| Extracted_Log | 12 cases<br>136 activity instances | Merge sub-activities | 12 cases<br>107 activity instances |
| True_Log | 12 cases<br>114 activity instances | None | 12 cases<br>114 activity instances |

Figure 6 depicts the results of the analysis in the form of two confusion matrices and contains the performance measures for the classification. The last row and the last column of the confusion matrices only display activity instances that have no match in the True_Log or in the Extracted_Log. Recall and precision are averaged by using the number of true instances for each activity class as the weight for the respective class in the calculation.

In total, the prototype correctly detected 92 activity instances, which corresponds to an accuracy of 69.70%. We noticed that only four observable activity instances in the True_Log had both an incorrectly matching and a correctly matching activity instance in the Extracted_Log. Therefore, the prototype predictions related to an observable activity instance could be regarded as very constant. Furthermore, the manual video data examination revealed that the prototype correctly detected six activity instances that were not contained in the original video annotations. These very brief activity instances were counted as correct and demonstrated the prototype's usefulness.

The average recall of 69.70% was mainly attributable to the prototype's failure to observe 25 activity instances included in the True_Log. As displayed in Figure 6, the prototype detected all activity instances for the activity classes "spread" and "stir," both of which involve visually prominent objects (i.e., pan and wooden spoon). It achieved high recall values for the activity classes "cut" and "pour," which also required visually prominent objects, as well as the activity class "sprinkle." The lower recall values for the activity classes "fold," "grate," and "transfer" may be due to the fact that no visually prominent objects were used to perform these activities. The swiftly performed activity "flip" had a comparatively small range of motion and was the only activity, in which misclassification with other relevant activity classes accounted for the majority of errors.

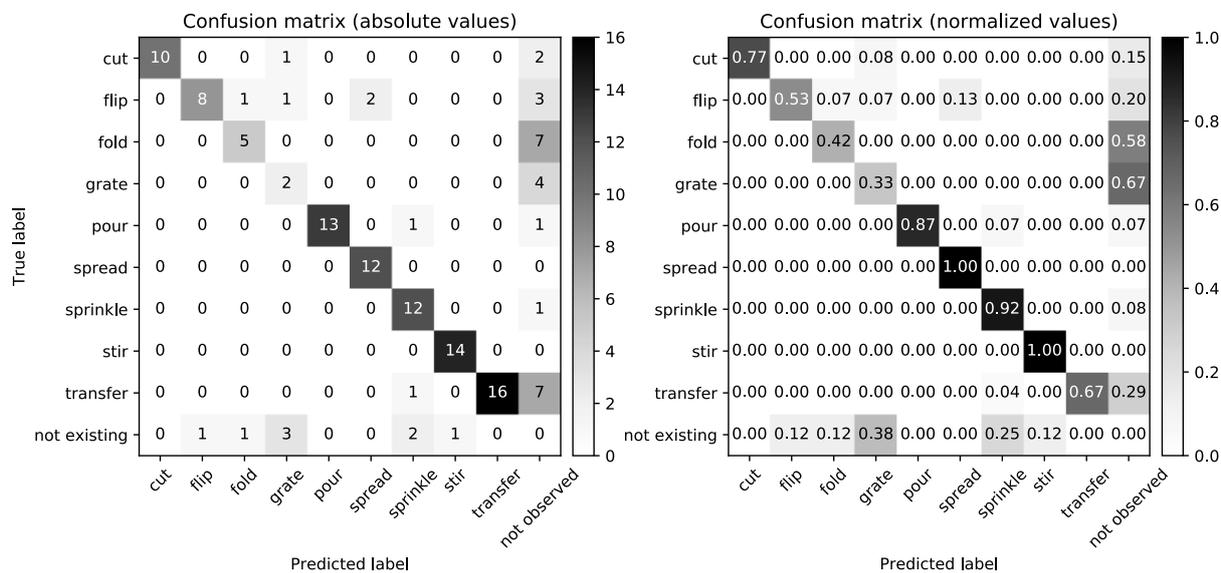

| | Performance Measures | | | | |
|---|---|---|---|---|---|
| Accuracy | 69.70% | Precision | 82.36% | Recall | 69.70% |

**Figure 6:** Confusion matrices comparing the activity instances detected by the prototype (Extracted_Log) with the observable behavior (True_Log)

In contrast, the average precision of 82.36% indicated that the prototype performed with a high level of accuracy once it detected an activity instance. The incorrect detections resulted from confusion about seven activity instances and the incorrect recognition of eight activity instances not recorded in the True_Log. With the exception of the activity class "grate," which again performed poorly, and the activity classes "fold" and "sprinkle," which achieved satisfactory precision values, the Extracted_Log showed an excellent level of precision. Interestingly, the detections for the activity class "transfer" were all correct, even though the prototype missed a high proportion of observable activity instances in the True_Log. The precision values support the hypothesis that the prototype was better at recognizing activities involving visually prominent objects.

Further worth noting is that the Activity Recognizer component seemed to perform significantly better for resources that are also included in its training data. For the three cases performed by an unknown resource, the prototype did not detect 12 observable activity instances and achieved considerably lower performance measures. This drop in performance may be attributed to the small amount of available training data. When only the nine cases with known resources were taken into account, an improvement in all performance measures was observed (i.e., accuracy 76.70%, precision 82.66%, and recall 76.70%).

## 8.3 Conformance Checking

After analyzing the extracted event log in detail, we performed conformance checking as a further means of evaluating the prototype's applicability and usefulness. To prepare the event log, we imported the True_Log and Extracted_Log in ProM. Since activities in the process model are strictly sequential, we filtered out all events with lifecycle:transition values "complete." We also applied the ProM plug-in "Merge subsequent events (AAABB -> AB)" to both logs in order to meaningfully aggregate all consecutive events in the same activity class. For conformance checking, we used the "Replay a Log on Petri Net for Conformance Analysis" plug-in that aligns event logs with process models [79]. Figure 7 visualizes conformance checking based on the prototype's detections contained in the Extracted_Log.

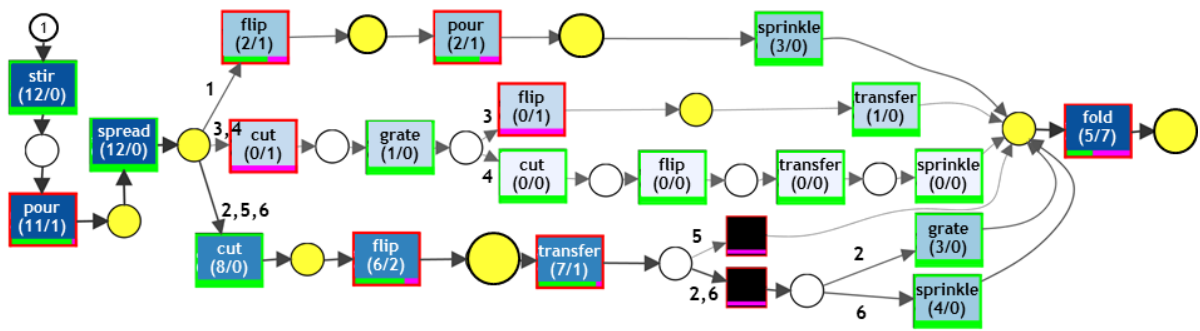

Figure 7: Conformance checking for the Extracted_Log (increased font for readability)

The plug-in counts how often log and model agree (i.e., synchronous moves, indicated by the left number below the activity name), how many activities along the selected path are not contained in the event log (i.e., move in the model, indicated by the right number below the activity name), and how many activities in the event log should not have happened according to the selected path through the model (i.e., move in the log, indicated by the yellow bubbles). Activities framed in red indicate asynchronous moves, and the bar at the bottom tracks the distribution of synchronous and asynchronous moves. Taking all of this into account, the approach computes a value for the fitness quality criterion, which indicates how much of the behavior contained in the event log is permitted by the process model.

The conformance checking of the True_Log resulted in a very high fitness of 97.98%, revealing that almost all behavior in the True_Log was in line with the process model [20]. Every trace fit its original process variant, and only one trace of the process variant "1) lemon sugar" contained five exclusive moves in the log. The inclusion of the prototype's six detected activities, which were missing in the original annotations, would have complemented the True_Log's real-world representation and reduced the calculated fitness value. Overall, the observable behavior for all 12 traces indicated a high level of compliance with the intended process model, which should have been reflected in the Extracted_Log.

For the Extracted_Log, the fitness value of 79.81% implied a high level of compliance with the process model, which resulted from 15 exclusive moves in the model and with 22 exclusive moves in the log. Due to the minority of erroneous activity instances in the Extracted_Log, only half of the selected paths for optimal alignments corresponded to the correct process variant. Three of the six differently aligned cases were performed by a resource that was not included in the training data. Although the alignment of all traces to their correct process variant would have reduced the fitness value of the Extracted_Log, the prototype successfully captured most of the observable behavior. The comparison of the two conformance checks emphasized the prototype's functional usefulness and its ability to generate event logs that represent reality and conform to the underlying process model.

### 8.4 Process Discovery

In the second process mining use case, we took the Extracted_Log as input to discover process models, while the True_Log served as a benchmark to assess the quality of the discovered models. In the following, we differentiate between the True_Log with the 12 evaluation cases (True_Log$_{12}$) and the True_Log with all 53 cases from the dataset (True_Log$_{53}$). Any discovered process model should be representative of both. In general, numerous process discovery techniques exist that address the trade-off between the four quality criteria for process models: fitness, precision, generalization, and simplicity [20]. For the purpose of this study, we selected the "Inductive Miner - infrequent (IMf)" variant [80] of the ProM plug-in "Mine Petri net with Inductive Miner" as this generates sound process models and filters infrequent behavior.

As for preprocessing, we removed the events with lifecycle:transition values "complete" and merged all subsequent events. To test how the separation of frequent and infrequent behavior affected the per-

accepted manuscript at Decision Support Systems                 24

formance, we generated multiple process models based on the Extracted_Log by increasing the separation threshold parameter. We computed precision (ProM plug-in "Check Precision based on Align-ETConformance" [81] using all optimal alignments) as well as fitness (ProM plug-in "Replay a Log on Petri Net for Conformance Analysis") based on the True_Log$_{12}$ and the True_Log$_{53}$. Supplementary material D provides a detailed report of the process discovery results.

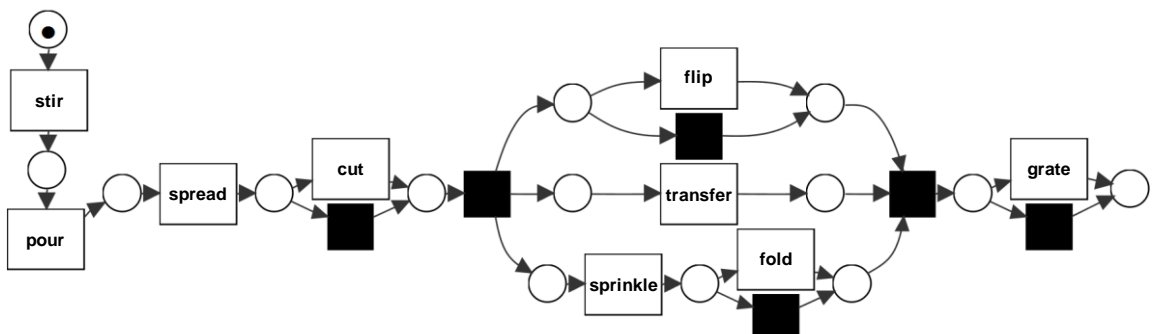

Figure 8: Process model 6, discovered from Extracted_Log (increased font for readability)

Since model 6 was discovered to be the best performing process model, it is illustrated as a Petri net in Figure 8. The first three activities of "stir," "pour," and "spread" are in the same sequential order as in the original process model However, the activity "fold" is not modeled as the final activity for all paths because the prototype did not always detect it correctly. Compared to the original model, the discovered model had fewer control-structures, resulting from the low amount of 12 traces in the Extracted_Log. Nevertheless, the process variant "6) goat's cheese spinach nutmeg" fit perfectly without any additional alignment. The generated process model satisfactorily balanced the trade-off between the four quality criteria and served as proof of the prototype's usefulness. The evaluations with 12 and 53 cases confirmed the validity of the discovered process model. We successfully minimized the negative impact of incorrect detections by filtering infrequent behavior during discovery, which would appear to be a promising approach for discovering process models based on partially erroneous event logs.

Overall, EVAL4 demonstrated the applicability and usefulness of the prototype and the ViProMiRA. Although the prototype was not able to entirely observe the reality, promising results regarding typical process mining measures emphasize substantial potentials for various process mining use cases.

# 9 Conclusion

## 9.1 Summary and Contribution

Since the volume of unstructured data is expected to increase further, and rapidly so, the importance of novel solutions for extracting structured insights from unstructured data and integrating these insights into process mining increases. This study takes a necessary step towards the reduction of blind spots in process analyses by answering the research question of how video data can be systematically exploited to support process mining. To this end, we have developed and thoroughly evaluated a reference architecture that bridges the gap between process mining and computer vision.

In our DSR process, we derived literature-backed DOs that synthesize the technical and business-related requirements to use video data for process mining. Based on these DOs, we aggregated existing knowledge from both computer vision and process mining to develop our artifact, the ViProMiRA. In doing so, we considered well-established methods for building and evaluating sound reference architectures. Since the computer vision and process mining capabilities integrated into the ViProMiRA are presented independently of their technical implementation, the ViProMiRA can adapt to technological advances. We performed multiple evaluation activities throughout our research process. First, we justified the identified research gap by analyzing existing approaches that combine process mining and video data (EVAL1 in Section 4). We also discussed the design specification of the ViProMiRA against the defined DOs and evaluated its understandability and real-world fidelity with fellow researchers (EVAL2 in Section 6.2). We further implemented the ViProMiRA as a software prototype in its most extensive instantiation variant (EVAL3 in Section 7). The Python prototype was based on open-source packages related to state-of-the-art computer vision approaches. To prove the ViProMiRA's real-world applicability and usefulness, we applied the prototype to a publicly available real-world video dataset that captured a manual process with concurrently running cases and multiple resources (EVAL4 in Section 8). After a detailed analysis of the event log that the prototype had extracted, we carried out two of the original process mining use cases (i.e., conformance checking and process discovery). As our results indicate, the proposed ViProMiRA is capable of extracting a high degree of process-relevant events from unstructured video data. The straightforward import of the gained log into ProM and the high discovery and conformance measures further demonstrate that ViProMiRA is equipped with useful and internally consistent event abstraction capabilities.

Our research makes a valuable contribution to the knowledge on extracting unstructured video data for process mining. By providing a reference architecture that spans both the computer vision and the process mining domains, our work lays a theoretical foundation for the enhancement of evidence-based decision support systems through video-based process mining. To the best of our knowledge, the ViProMiRA is the first approach to systematically exploit video data for process mining. Furthermore, our prototypical implementation is the first technical artifact that enables the unobtrusive extraction of concurrently running cases performed by multiple actors. As an additional contribution, we also facilitate the future implementation of such a system by providing an end-to-end example of how to use open-source frameworks to most efficiently instantiate the proposed ViProMiRA components.

### 9.2 Limitations and Future Research

While validating the ViProMiRA, we also identified some limitations, which we discuss in the same order as our reference architecture design and evaluation. Firstly, to allow for flexibility regarding the design of the ViProMiRA, we decided to use a semi-concrete level of description. Although the prototypical instantiation presented in Section 7 can serve as a blueprint, the ViProMiRA itself cannot serve as a fully specified implementation concept as it is not directly transferrable to practical use. Secondly, video data contains potentially sensitive information. Thus, strict compliance with regulations such as Europe's General Data Protection Regulation as well as other privacy and transparency imperatives is required [14]. To ensure this compliance, the design of our modular ViProMiRA implements strategies to preserve privacy while extracting process information from video data. For instance, blind sensors that apply filters to video material before it is stored can be used as a data sources. While blind sensors make it possible to avoid the identification of people, computer vision capabilities can still be applied with a reduced level of accuracy [69]. Thirdly, the concept of using computer vision approaches to extract information from video data requires sufficiently advanced optical sensors and resources for the underlying deep learning models. As the quality of available cameras, hardware, and algorithms is increasing at a rapid rate, the impact of this technical limitation should be reduced in the near future.

Fourthly, video data can only partially fill in blind spots and should therefore be enriched with further data, such as process data from core systems or other contextual data that could provide an end-to-end process perspective. Fifthly, our prototypical instantiation is entirely based on supervised learning. Therefore, it restricts the extractable information to predefined classes and is dependent on labeled training data. Finally, regarding the evaluation of the ViProMiRA, we have to point out that the Crêpe Dataset does not represent a real-world process and only contains 90 minutes of video data. Due to the rather simple model and well-annotated video, however, this dataset was well-suited to evaluate the technical feasibility of the ViProMiRA and illustrate the results in a clear and comprehensive manner. Thus, although EVAL4 allowed us to assess the artifact's usefulness when it comes to process mining analysis (i.e., conformance checking or discovery), feedback from actual process mining experts and an application in a real-world setting would have strengthened the evaluation even further.

To address these limitations and explore novel concepts for leveraging unstructured data in process mining, we have identified multiple avenues for future research. Firstly, future researchers should examine unsupervised approaches for implementing information extraction components. For instance, unsupervised learning might be capable of extracting basic activities that can be aggregated to specific activities (e.g., stirring batter is a hierarchical arrangement of grasping, moving the arm, and releasing). Furthermore, unsupervised methods may identify additional activity classes that are not considered in a supervised scenario. Their inclusion would facilitate a more holistic representation of real-world scenarios. Secondly, future research should also use larger, process-related datasets that provide high-quality spatio-temporal labeling of continuous video data that contains more activity classes. Thirdly, the potential of blind sensors that continuously comply with data protection laws should be investigated in detail. Fourthly, the analysis of more complex processes should also be examined. Fifthly, to further reduce the number of blind spots and enable an end-to-end process view, future research should build on ViProMiRA and enhance it with further valuable data sources, for instance, by considering audio files as well as sensor data and text data. Lastly, the ViProMiRA builds on existing concepts for computer vision and process mining. Since new concepts and approaches will be developed in both disciplines, these should be continuously reviewed in order to update the ViProMiRA.

accepted manuscript at Decision Support Systems 28